# FITS: Towards an AI-Driven Fashion Information Tool for Sustainability

**Daphne Theodorakopoulos[a,*,1], Elisabeth Eberling[b,c], Miriam Bodenheimer[b], Sabine Loos[d] and Frederic Stahl[a]**

[a]Marine Perception Research Department, German Research Center for Artificial Intelligence (DFKI)
[b]Competence Center Sustainability and Infrastructure Systems, Fraunhofer ISI
[c]Faculty of Economics and Management, TU Berlin
[d]Center for Responsible Research and Innovation, Fraunhofer IAO

**Abstract.** Access to credible sustainability information in the fashion industry remains limited and challenging to interpret, despite growing public and regulatory demands for transparency. General-purpose language models often lack domain-specific knowledge and tend to "hallucinate", which is particularly harmful for fields where factual correctness is crucial. This work explores how Natural Language Processing (NLP) techniques can be applied to classify sustainability data for fashion brands, thereby addressing the scarcity of credible and accessible information in this domain. We present a prototype Fashion Information Tool for Sustainability (FITS), a transformer-based system that extracts and classifies sustainability information from credible, unstructured text sources: NGO reports and scientific publications. Several BERT-based language models, including models pretrained on scientific and climate-specific data, are fine-tuned on our curated corpus using a domain-specific classification schema, with hyperparameters optimized via Bayesian optimization. FITS allows users to search for relevant data, analyze their own data, and explore the information via an interactive interface. We evaluated FITS in two focus groups of potential users concerning usability, visual design, content clarity, possible use cases, and desired features. Our results highlight the value of domain-adapted NLP in promoting informed decision-making and emphasize the broader potential of AI applications in addressing climate-related challenges. Finally, this work provides a valuable dataset, the SustainableTextileCorpus, along with a methodology for future updates. Code available at https://github.com/daphne12345/FITS

## 1 Introduction

The fast-evolving fashion sector demands that retailers react rapidly to new trends, quick production cycles, and low-cost next-day deliveries, often at the expense of poor labor conditions and environmental exploitation. Certifications are one possible solution; however, they often lack transparency, supporting evidence, or verification [6] or are selectively used for marketing, raising concerns about greenwashing [25]. Additionally, the vast amount and the fact that even trusted certifications tend to focus on specific aspects of sustainability (some prioritize social issues like fair wages, while others emphasize ecological factors such as carbon emissions) make it difficult for consumers to get an overview. This leads to a lack of comprehensive, accessible, and independent sustainability data, making it challenging to report holistically on the sustainability of textile brands [9].

At the same time, recent advances in Natural Language Processing (NLP) have focused heavily on general-purpose language models, which, despite their versatility, often generate incorrect or fabricated information – a phenomenon known as "hallucination" [13]. This poses a critical problem for domains like sustainability reporting, where factual correctness and source credibility are essential. Moreover, large language models are very energy-intensive, and domain-specific, fine-tuned models offer a more lightweight, accurate, and environmentally friendly alternative for tasks requiring domain expertise and trustworthiness. Additionally, previous work has shown that these smaller models, in this case BERT-based models, outperform large language models for climate-change data [10].

To approach this, we propose the Fashion Information Tool for Sustainability (FITS). Our system classifies credible, unstructured sustainability information about fashion brands gathered in a self-curated dataset using fine-tuned BERT models, which are a class of pre-trained language models [5]. To evaluate our algorithm, we showcase the model within a tool and discuss it in focus groups. To our knowledge, no existing system uses fine-tuned transformer models to classify credible sources for sustainability in the fashion sector.

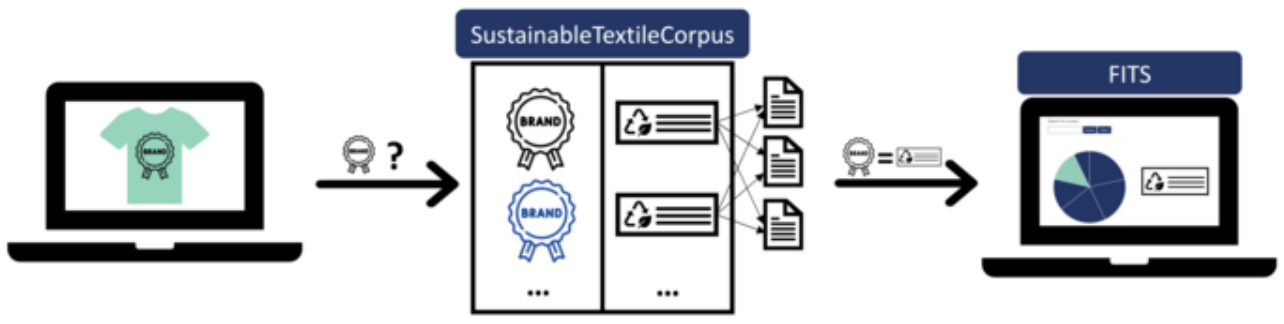

**Figure 1.** The Figure shows the algorithm for extracting and classifying sustainability information. Users can search the SustainableTextileCorpus by brand, and FITS returns and classifies relevant sustainability data, which users can explore interactively.

We developed and evaluated FITS as a prototype (see Figure 1). It allows users to search and explore brand-level sustainability data through an intuitive interface. Unlike existing solutions that rely on certifications, FITS aggregates independent insights from NGO reports and scientific publications. Our focus is specifically on sustainability insights beyond certifications, ensuring credibility by sourcing data from well-recognized NGOs in the textile sector and peer-reviewed scientific publications. This leads to our research question: How can AI techniques be applied to provide credible information on the sustainability of textile brands effectively?

* Corresponding Author. Email: d.theodorakopoulos@ai.uni-hannover.de.
1 Currently working at AI Institute, Leibniz University Hannover.

Our methodology includes creating the SustainableTextileCorpus by collecting relevant texts, defining sustainability classes (e.g., "wages & hours"), and labeling part of the data (Section 3). We developed an NLP algorithm that fine-tunes a BERT model (Section 4). We defined the task as a multi-label classification, as text snippets often belong to multiple classes. Hyperparameter Optimization (HPO) was applied to refine model performance, especially important given the small dataset. We then defined FITS requirements and implemented the tool accordingly (Section 5). Finally, we evaluated the tool within two focus groups (Section 6).

In summary, our contributions are fourfold: (1) FITS, a prototype system that classifies sustainability information from credible text sources using fine-tuned BERT models, a low-compute solution suitable for website integration by online retailers; (2) a multi-label classification scheme to support structured sustainability reporting in the textile sector; (3) the SustainableTextileCorpus, a curated dataset with an update methodology; and (4) insights from focus group evaluations to guide future development. This work shows how domain-adapted NLP can avoid hallucinations, reduce compute costs, and support credible, AI-assisted sustainability action.

## 2 Related Work

The **sustainability challenges in the textile sector** are concerning, given the industry's environmental and social footprint. The textile sector is responsible for around 20% of global clean water pollution [8]. Additionally, in the EU, it generated 121 million tonnes of greenhouse gases in 2020 [7]. Moreover, the sector contributes to extensive waste generation (with less than 1% being recycled into new fibers globally) [7]. Labor issues are also prevalent, with many supply chains linked to inadequate working conditions. Despite consumers wanting to be informed about a brand's sustainability [6], purchasing decisions remain largely unaffected due to barriers such as lack of knowledge and limited transparency [21, 14, 12]. Therefore, companies should focus on educating and engaging consumers to drive sustainable consumption [26]. In response, many brands now pursue eco-friendly materials and certifications to ensure ethical practices. This shift reflects a push for transparency and responsible sourcing in fashion, aligning industry goals with consumer demand for sustainable products [28]. However, navigating these can be complex for consumers, and not all certificates can be trusted [25].

Machine Learning (ML) has demonstrated its effectiveness in tackling climate-related challenges, as highlighted by Rolnick et al. [30]. **NLP** plays an essential role in sustainability research by facilitating the analysis of unstructured data. For example, Webersinke et al. [36] created ClimateBERT, a language model trained on climate-related data. Pre-trained BERT models excel in various NLP tasks [5], including the assessment of companies' ESG criteria [33]. Similarly, Stammbach et al. [34] proposed using NLP for detecting environmental claims made by companies. Recent studies suggest that BERT-based models remain more efficient than large language models for text classification tasks in the climate domain [10], making it reasonable to use BERT-based models like in our study, thereby saving energy due to their significantly smaller model size.

**AI Techniques in the textile sector** are transforming processes to boost efficiency. In manufacturing, fabric defects are detected, reducing material waste [16]. Predictive models improve demand forecasting, mitigating overproduction and inventory shortfalls [35], while AI-driven supply-chain optimizations enhance resource allocation [24]. In e-commerce, AI has transformed the customer experience [27] with personalized shopping through recommender systems or AI-driven virtual try-ons [11]. Moreover, data analytics help optimize business strategies and customer engagement [1].

In recent years, efforts have been made to use **AI-driven sustainability in the textile sector**. In their review, Ramos et al. [29] found that AI in the sustainable fashion industry primarily targets the business and customer sectors; they criticize that no articles targeted the governmental sector. According to them, AI tools enhance garment manufacturing, design, and handling processes. For customers, the focus is on equipping users with tools to encourage the purchase of sustainable products [29], aligning closely with the objectives of FITS. For example, "Good On You"[2] rates fashion brands on their sustainability. Related work also uses text processing to retrieve sustainability information; for example, Li and Zhao [17] use CSR reports, and Satinet and Fouss [32] use life cycle assessment data. Notably, scientific articles and NGO reports have not previously been employed as data sources. To fill this gap, we develop FITS.

## 3 Creation of the SustainableTextileCorpus

The SustainableTextileCorpus is a self-compiled dataset containing sustainability-related texts about the fashion industry, drawn from publicly available sources such as scientific publications and NGO reports. The corpus was constructed by collecting relevant articles using predefined keyword lists, cleaning, and annotating the data. The section concludes with a quick data analysis to show the relevance of the collected data to the task. The dataset serves as the foundation for training our ML models and building FITS. The dataset is unpublished as brand sustainability data changes quickly and may become outdated or misleading. To avoid spreading potentially incorrect claims, especially in a sensitive domain, we instead provide detailed reconstruction steps and the full list of sources used (all of them open-source).[3] For future use, rebuild the corpus with the most recent articles following our steps.

**Keyword Lists.** To ensure relevance, two keyword lists were curated by sustainability experts.[4] The experts are both researchers specialized in the social and ecological sustainability of supply chains, focusing on the fashion sector. The first list (the brand list) contains textile brands relevant to the European market. The second list (the keyword list) contains keywords corresponding to 19 expert-defined sustainability-related issues, later used as sustainability classes for the text classification. The starting point for the keyword list was the compilation of sustainability-relevant criteria from the German website "Siegelklarheit".[5] The list was enhanced based on a literature review and industry standards. Eleven issues represent social aspects of sustainability, and the other eight represent environmental factors.

**Data Collection.** The focus explicitly lies on information beyond sustainability certifications. To ensure the data's credibility, we only considered (1) scientific publications and, additionally, (2) reports of five selected NGOs.[6] Articles were considered if they contained specific keywords and were written in English because we did not want to mix languages. Mixing languages often reduces accuracy and reliability because most models are designed to handle one language

[2] https://goodonyou.eco/
[3] The literature list forming the corpus is in the GitHub Repository.
[4] See the GitHub Repository for both lists.
[5] https://www.siegelklarheit.de/
[6] "Clean Clothes Campaign", "SOMO", "Asia Floor Wage Alliance", "Global Labour Justice" and "Labour Behind the Label"

at a time. The scientific publications were searched via Scopus[7] and Web of Science[8], where the title or abstract must contain "textile", "garment", "clothing", or "clothes" in addition to "supply chain" or "sustainability". Only articles from 2017 until October 24, 2022 (data collection date) were kept. The titles and abstracts were filtered with the list of textile brands. We only kept entries with a DOI to ensure we can reference the data source. If no DOI was given, we tried to find it with the Crossref API.[9] Next to the abstracts, we also downloaded the available complete texts if they had a free license.[10] The licenses were also extracted based on the DOI using the Crossref entries. This resulted in 411 full-text articles. The NGO reports were automatically downloaded from the respective download page. We collected all sublinks on these pages and then again all sublinks of the subpages, which we repeated a third time. Subsequently, the links were filtered to end either with ".pdf" or "somo.nl/download". These links were downloaded automatically. This resulted in 971 NGO reports. We identified 3068 abstracts and 1382 full texts from scientific publications and NGO reports.

**Pre-processing.** The raw text was extracted from all downloaded PDFs. Empty, duplicate, and non-English entries were removed from both the abstracts and the texts from the PDFs. Next, the text was cleaned by eliminating line breaks, extra spaces, extra punctuation, and the text following copyright symbols. Texts shorter than 100 characters were discarded, as they most likely do not contain meaningful content. Subsequently, each text was tokenized into sentences, which must contain a verb and be more than three words long. After that, a sliding window of three sentences was created for context with an overlap. That means each sentence was once in the center. We chose three sentences because, in a previous study[11], participants found this to provide an appropriate level of information depth. As transformer models can only handle a fixed context length, text snippets over 1000 tokens were disregarded. The data still contained noise, so we only kept a text passage if it included a word from each, the brand list, and the keyword list. The keywords contain British and American spellings and singular and plural forms. The final dataset contains 5,093 three-sentence text passages.

**Labeling Process.** A subset of the three-sentence texts was manually labeled by the two domain experts using the predefined sustainability classes, allowing for multiple labels per text. Initially, we applied keyword-matching (see baselines in Section 4.1) and sampled at least two texts per class, up to the proportional sample size based on its occurrence in the dataset, to ensure representation. For an efficient human labeling process, we used LightTag.[12] As an additional class, the text passage could be labeled as irrelevant. Finally, the annotations were downloaded and prepared for model training. The final dataset has 582 labeled samples.

**Analysis of the SustainableTextileCorpus.** We looked at the most frequent noun phrases in the SustainableTextileCorpus to understand if the data is relevant to the task (see the word cloud in Figure 2). The larger the words appear, the more frequent they are. The word cloud shows that the extracted texts are relevant for sustainability in the fashion industry, as they include thematically appropriate keywords such as "Paris agreement", "Bangladesh" or "garment workers". Additionally, big fashion brands and retailers, such as "Amazon" or "Adidas" are very frequent. It indicates a bias in the data toward well-known brands, as smaller brands have fewer samples. This outcome was expected, as scientific publications often focus on prominent brands. The class distribution is displayed in Figure 3. The dataset appears to be imbalanced. While some classes are rare or absent, i.e. "vegan", others, like "wages & hours", are over-represented. This imbalance, combined with the limited data, likely affects the text classification model's performance and constrains the quality of FITS.

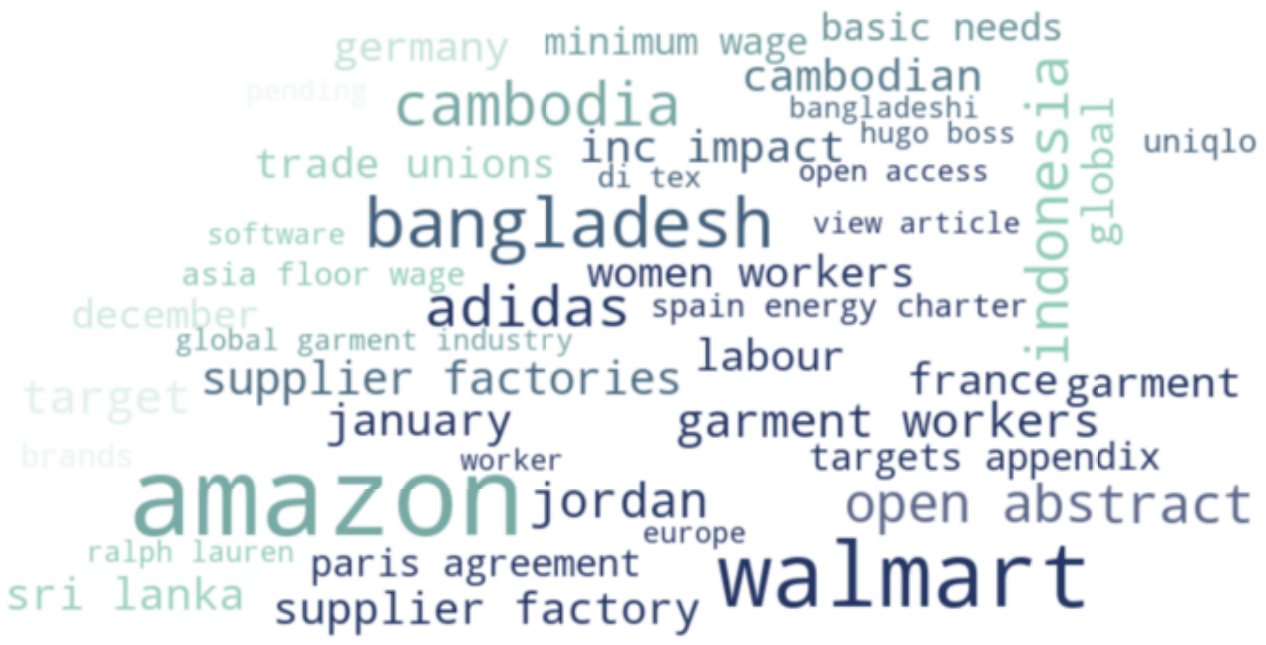

**Figure 2.** Word cloud of the most frequent noun phrases in the SustainableTextileCorpus.

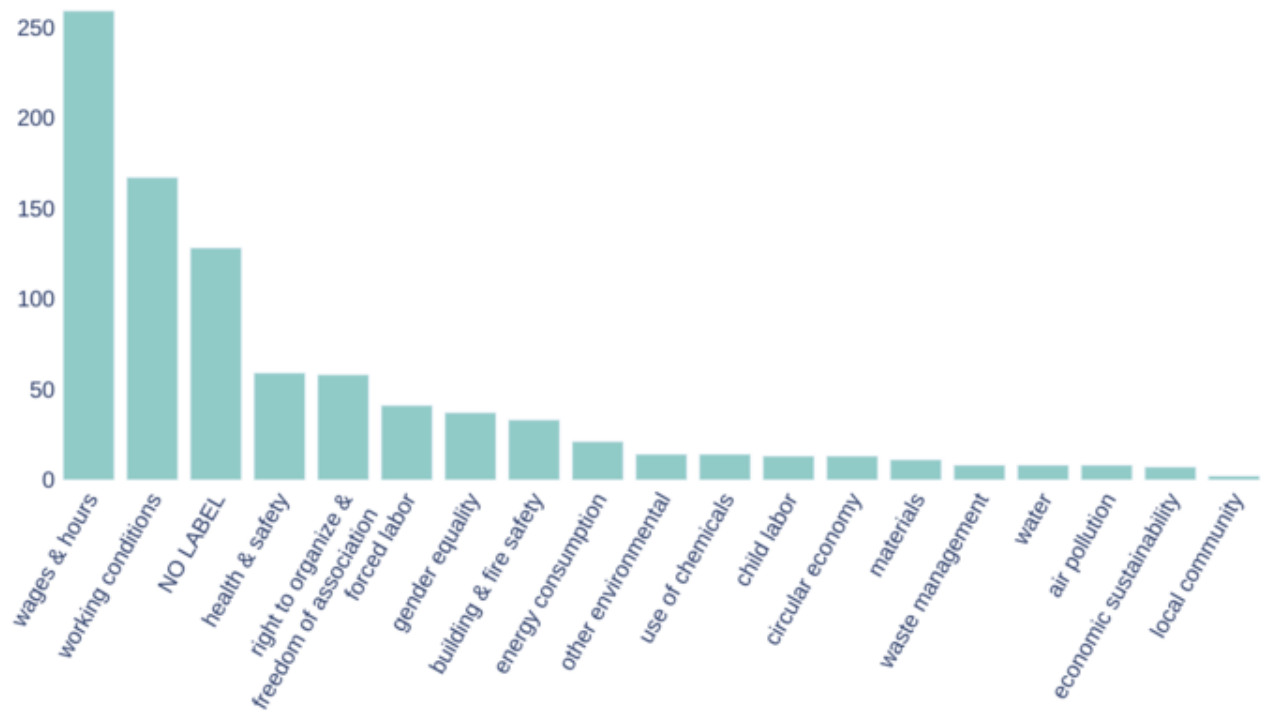

**Figure 3.** Class distribution of the labeled dataset.

## 4 Model Training and Optimization

We developed a multi-label text classification model to categorize sustainability-related texts from the SustainableTextileCorpus. Fine-tuning pre-trained language models was central to our approach, as they perform well even with limited labeled data [5]. Moreover, we performed HPO to boost performance further. We compared different BERT-based models against each other and also against three baselines: a keyword-matching baseline, a statistical ML approach, and a BERT base model without HPO.

### 4.1 Model Training

We used Huggingface's transformer library [37] for model training. We fine-tuned five pre-trained models with our data: BERT base [5], RoBERTa base [19], DistilBERT [31], SciBERT [3], and Climate-BERT [36]. We decided on BERT and RoBERTa because they are widely used and robust. We chose DistilBERT as a resource-efficient alternative. Moreover, we included SciBERT because it is pre-trained with scientific texts, relevant to much of our corpus. Finally, ClimateBERT was chosen as it is pre-trained on climate-related texts, including research data. We used the cased versions of the models to avoid losing the meaning of synonyms and the base version because the large version can more easily overfit the relatively small dataset.

[7] https://scopus.com
[8] https://webofscience.com/
[9] https://crossref.org/
[10] Included licenses: CC BY 3.0, CC BY 4.0, CC BY-NC 3.0, CC BY-NC 4.0, CC BY-SA 4.0, CC BY-NC-SA 4.0
[11] https://www.zusina-guide.de/glaubwuerdige-nachhaltigkeitsinformation/#Wie%20viel
[12] By now, primer.ai acquired LightTag.

**Table 1.** Performance of the best HP configuration on the test set (mean and standard deviation across seeds).

| Model | Weighted F1-score | Macro F1-score | Micro F1-score | Weighted precision | Weighted recall |
|---|---|---|---|---|---|
| Keyword Matching | 0.58 | 0.50 | 0.63 | 0.61 | 0.64 |
| TF-IDF + SVM | $0.545 \pm 0.004$ | $0.343 \pm 0.004$ | $0.582 \pm 0.005$ | $0.667 \pm 0.002$ | $0.495 \pm 0.006$ |
| BERT base (no HPO, 20 epochs) | $0.627 \pm 0.014$ | $0.368 \pm 0.029$ | $0.674 \pm 0.01$ | $0.629 \pm 0.018$ | $0.661 \pm 0.014$ |
| BERT base | $0.66 \pm 0.016$ | $0.439 \pm 0.034$ | $0.696 \pm 0.012$ | **0.673** $\pm 0.02$ | $0.687 \pm 0.024$ |
| RoBERTa base | **0.683** $\pm 0.017$ | **0.479** $\pm 0.046$ | **0.705** $\pm 0.014$ | $0.665 \pm 0.023$ | **0.734** $\pm 0.022$ |
| DistilBERT | $0.644 \pm 0.006$ | $0.426 \pm 0.013$ | $0.675 \pm 0.005$ | $0.652 \pm 0.014$ | $0.669 \pm 0.001$ |
| SciBERT | $0.648 \pm 0.02$ | $0.451 \pm 0.037$ | $0.675 \pm 0.017$ | $0.661 \pm 0.031$ | $0.68 \pm 0.014$ |
| ClimateBERT | $0.652 \pm 0.018$ | $0.437 \pm 0.024$ | $0.679 \pm 0.013$ | $0.666 \pm 0.027$ | $0.678 \pm 0.028$ |

**Procedure.** Before the training started, the data was shuffled and split into training (70%) and test (30%) sets. Subsequently, the training set was again divided into training (80%) and validation (20%) sets. Next, the data was tokenized, and embeddings were generated with the pre-trained model. The data was truncated if it exceeded the maximum character length of the model. The model was fine-tuned on the training data. The training was configured to adjust the batch size automatically and used label smoothing, weight decay, a learning rate scheduler, and learning rate warm-up to stabilize and regularize training. These hyperparameters (HPs), as well as the learning rate and the training epochs, were tuned. We used AdamW [20] as an optimizer. As a loss function, we chose binary cross-entropy with logits, which is well-suited for multi-label tasks where each instance can belong to multiple categories.

**Evaluation.** For evaluation, the model predictions were passed through a sigmoid activation to obtain probabilities, which were then thresholded to generate binary predictions. That means each class was considered a prediction if the model's probability surpassed a certain threshold, which was also subject to HPO. Therefore, it is also possible that a text passage does not belong to any class. Since the dataset is imbalanced, the F1-score is our primary performance metric, calculated using the micro, macro, and weighted averages.

**Baselines.** Three baselines were implemented. The first one is a simple keyword-matching approach. Using the list of sustainability keywords sorted by issues, we assigned a text passage to a class if a keyword matched, partially leading to multiple labels. This process was applied to the entire test set and compared with the human-assigned labels. The second baseline is a statistical approach that trained a Support Vector Machine (SVM) based on Term Frequency-Inverse Document Frequency (TF-IDF) scores. We chose SVMs because they can handle high-dimensional, sparse data well. The TF-IDF score is high if an n-gram (a sequence of n words) is frequent in a few text passages but rare in the rest. This reflects important terms for specific classes. We excluded stop words, and the maximum length of the n-grams was subject to HPO. The output was passed to a one-vs-all classifier, which trains one SVM per class to realize multi-label classification. The SVMs had a linear kernel, where the regularization HP C is tuned. The third baseline is the BERT base model trained for 20 epochs with the default configuration provided by the transformer library. This served only as a baseline for HPO.

**Hyperparameter Optimization.** For both the SVM baseline and the BERT models, HPO is performed to explore a range of potential HP values and find the combination that yields the best model performance. This is especially important since the amount of labeled data is scarce, and multi-label classification is complex. HPO was performed via the SMAC library [18]. SMAC is a Python library that uses Bayesian Optimization to find the best HP configuration for an algorithm. We used the HPO facade, which uses a random forest as a surrogate model, and log expected improvement [2] as an acquisition function. Table 2 shows the HP space for the BERT models. For the statistical baseline, we only tuned the maximum n-gram length from 1 to 4 as an integer and the SVM HP C as a float value from 0.1 to 10. C regulates the trade-off between the hyperplane margin and misclassification. Each optimization was run for 1000 trials for five different random seeds with the weighted F1-score on the validation data as the optimization metric. After that, we retrained the model with the best configuration per seed and evaluated it on the test set.

**Table 2.** HP search space for the BERT models.

| HP | Range/Choices | Type |
|---|---|---|
| Learning Rate | $[1 \times 10^{-6}, 0.01]$ | Logarithmic |
| Weight Decay | $[0.0001, 0.3]$ | Logarithmic |
| LR Scheduler Type | linear, cosine, cosine with restarts, polynomial, constant, constant with warmup, inverse sqrt, reduce lr on plateau | Categorical |
| Warmup Ratio | $[0.0001, 0.1]$ | Logarithmic |
| Label Smoothing | $[0.0001, 0.1]$ | Logarithmic |
| Threshold | $[0.3, 0.6]$ | Linear with a step size of 0.01 |
| Epochs | $[15, 35]$ | Linear integer |

## 4.2 *Results and Discussion*

Table 1 shows the test performance for the baselines and the different fine-tuned models of the respective best HP configuration (mean and standard deviation across seeds). Our best-performing fine-tuned BERT model is RoBERTa, with an average weighted F1-score of 0.683, outperforming the baselines. For all models, the scores are far from 1, which is due to the small dataset, the imbalance of the data, and multi-label classification being a difficult task. Further dataset expansion could potentially enhance performance.

**Table 3.** Hyperparameter configuration for the best RoBERTa Model.

| Learning Rate | Weight Decay | LR Scheduler | Warmup Ratio | Label Smoothing | Threshold | Epochs |
|---|---|---|---|---|---|---|
| 6.9e-05 | 0.02627 | cosine | 0.04439 | 0.00136 | 0.33 | 34 |

While keyword-matching and SVMs are strong baselines, the BERT-based models are better suited for the task. Keyword-matching performed well but struggled with false positives, affecting precision and recall. The SVM baseline showed a lower recall, indicating it classified fewer instances, which led to worse performance than the BERT models. These results highlight the advantage of fine-tuning pre-trained models like BERT, which better capture the semantic meaning of the texts. The third baseline of BERT without HPO is slightly worse than the other BERT models, where the HPs were optimized. This shows the positive effect of HPO for the use case.

All F1-scores are highest for the tuned RoBERTa model, which is due to its high recall. However, the tuned BERT model has the highest precision and is closely followed by the other models. RoBERTa's outstanding performance makes sense as it is the refinement of BERT. Unfortunately, the DistillBERT model performed much worse; thus, it should not be used as a resource-efficient alternative here. Surprisingly, although the SciBERT and ClimateBERT models are pre-trained on a related domain, they performed worse than BERT. The micro-F1-score weighs all classes the same; considering that metric, the BERT model would be almost as good a choice as the RoBERTa model. The best RoBERTa model with seed two had the configuration displayed in Table 3. The low classification threshold (0.33) was particularly notable, reflecting the model's uncertainty and its strategy of classifying more samples to improve recall.

## 5 Tool Development

Based on the SustainableTextileCorpus and the classification algorithm, we implemented FITS, which assigns text passages to the sustainability classes and allows users to explore the data and search for keywords visually. Moreover, users can upload their own data. We formulated some meta-requirements that FITS should fulfill based on our expectations and practitioners' opinions, which we gained through a preceding workshop.[13] The list of requirements served as a benchmark to evaluate our tool, but it is not a complete list.

**R1:** The tool should only display credible information. Ideally, only verified information certified by third parties is displayed.
**R2:** The sources of the information should be traceable.
**R3:** The information should be as dense as possible.
**R4:** It should be possible to upload and analyze your own data.
**R5:** The tool should sort the text into certain categories.
**R6:** It should be possible to filter by brands.
**R7:** The tool should be interactive.
**R8:** The tool should be easily manageable.

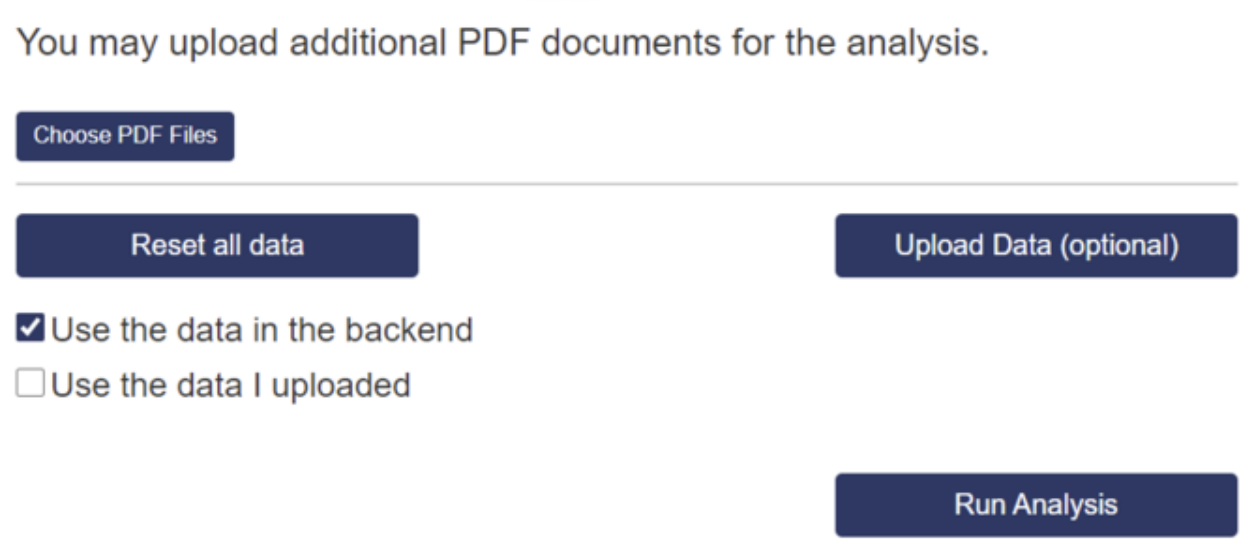

**Figure 4.** Landing page of FITS. The user can optionally upload data and select the data sources for the analysis.

Considering the requirements, FITS was implemented as a web-based Flask application[14] that can be deployed on a server. A user can upload additional data in the first window (Figure 4). Then, the user can decide on only the uploaded data and/or the data in the backend, i.e., the SustainableTextileCorpus. Subsequently, the data is preprocessed and classified as described in Sections 3 and 4. The following window (Figure 5) displays the result of the text classification as a pie chart and a table of texts. If a user clicks on a segment, only text passages that belong to that class are displayed. A full-text search has also been implemented, allowing users to display only text passages in the selected class that contain the searched keyword. Subsequently, the implementation will be explained in more detail.

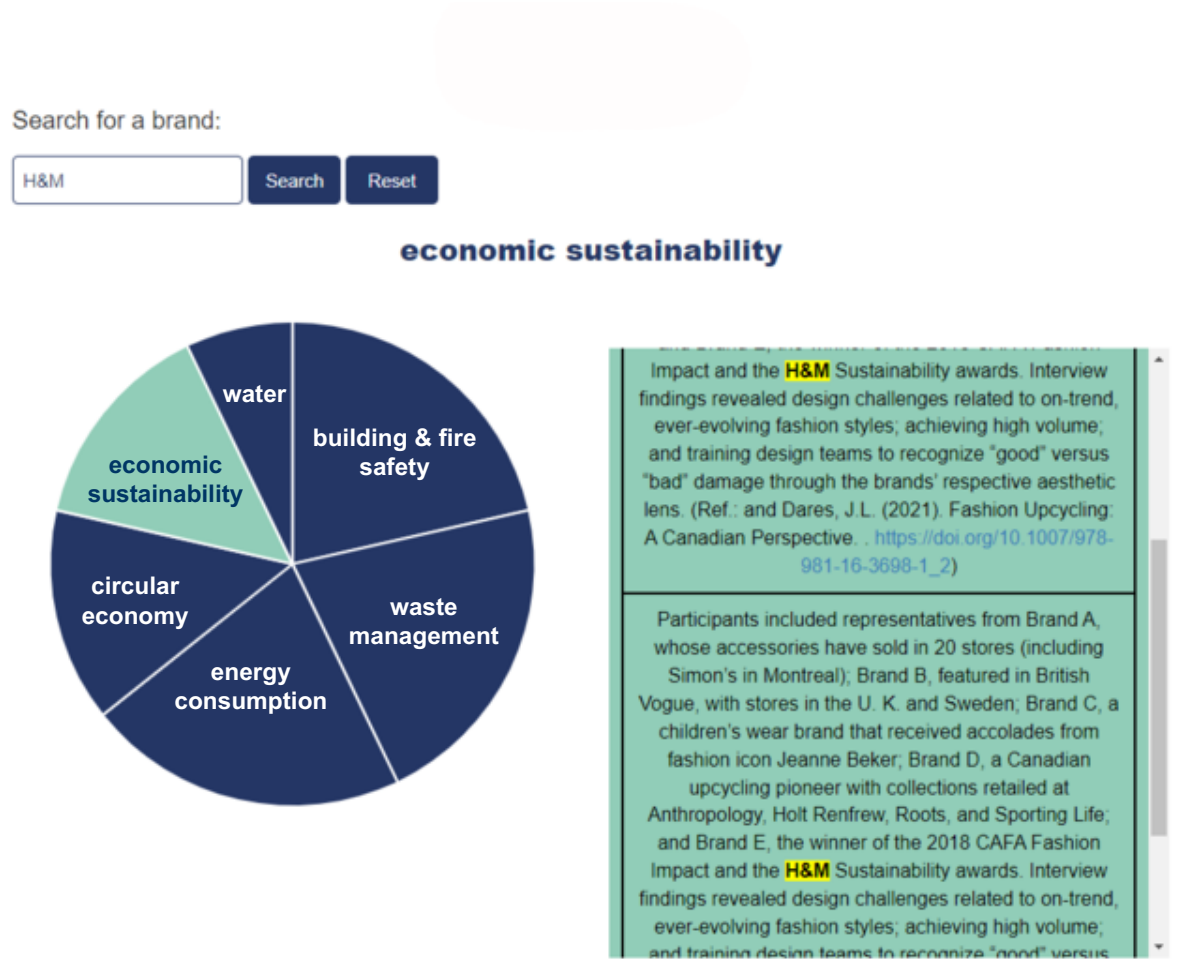

**Figure 5.** The data results page of FITS. The user can interact with the pie chart to view the texts within the classes and filter the results via the search bar. The example shows the results for "H&M" within the class economic sustainability (selected segment in lighter blue). The labels were only added in the paper for illustrative reasons and were not present in the tool.

**File and User Management.** As per R4, FITS supports the uploading of user-specific data. For example, some users might have internal information about a company, or new data may become available. Users can upload files through a simple interface where files are stored on the server under a user-specific session ID. This ensures that users cannot access other users' uploaded data. A cleanup function removes all user files when the session ends, maintaining data privacy and preventing unnecessary storage accumulation.

**Data Processing.** If new data was uploaded, it is preprocessed in the same way as described in Section 3. In addition, the resulting text passages are classified using the trained model. Only the texts with a class label are kept. If the data in the backend is selected, the datasets are merged. While the data is being processed, which only takes a few seconds to a few minutes depending on the amount of uploaded data, a loading page is displayed with a spinner. Since R2 requires providing references for the information, a hyperlink to the source document is appended to each text; it is generated from the DOI for scientific papers, while for NGO reports, it links to their websites. For uploaded files, the file name is displayed after the text passage.

**Interactive Visualization.** The processed dataset, whether created from uploaded files or the SustainableTextileCorpus, forms the basis for generating the interactive pie chart and table. The pie chart illustrates the distribution of topics within the dataset, with each class represented in the data by a segment. The pie chart dynamically adjusts colors to indicate the selected segment, improving data interpretation. Users can filter the data by a keyword search, which dynamically updates the displayed results. Only texts containing the keyword are kept, and the keyword matches are highlighted in yellow within the text. The design considers usability with features like support for the Enter key during the search, reset functionality, and informative feedback for empty or unmatched search results.

[13] The workshop was part of a bigger research project, where a prototype of FITS was shown, and some verbal feedback was collected. This was only a small part of the workshop, not the main focus.
[14] https://flask.palletsprojects.com

**Error Handling.** Error-handling mechanisms are integrated into the application. For example, the system informs the user if the uploaded data cannot be processed or lacks relevant content, i.e., was not classified by the model. It defaults to displaying existing backend data when appropriate. Moreover, alerts are triggered to give the user feedback for successful actions, such as uploading the data.

## 6 Qualitative Evaluation: Focus Groups

Two virtual focus group workshops were conducted to evaluate FITS. The following sections describe and discuss the method and setup, and present the feedback. Focus groups leverage group dynamics to uncover diverse opinions, shared concerns, and collective expectations [23, 15]. The interactive setting encourages participants to build on each other's insights. Additionally, the structured yet flexible nature ensures that all topics are addressed while leaving room for unexpected insights. Focus groups are well-suited to the fashion sector, where sustainability is tied to values and trends.

### *6.1 Experimental Setup*

Each workshop involved ten participants (20 in total), selected based on diversity in age, gender, and family status. All participants had to be regular online shoppers who are technologically proficient and proficient in English and German, since the workshops were held in German and FITS is in English. An entry question revealed that some participants found it easier to shop sustainably online, while others faced challenges distinguishing sustainable options.

Before the workshop, the participants received two PDF documents. A collaborative online whiteboard (Conceptboard) was used to gather real-time annotations. FITS was hosted locally and accessed through a provided URL. A structured questionnaire guided the discussion, covering usability, visual design, content clarity, potential use cases, and desired features. The data was collected via recordings with transcripts, Conceptboard inputs, and notes. All inputs were merged into session summaries and analyzed using a summarizing content analysis [22]. The material was reduced to core statements to identify themes, which will be summarized in the next section. Each session lasted two hours and followed this format:

1. **Introductions**: The session started with a warm-up question regarding the challenges participants faced when searching for sustainable products online.
2. **Research Overview**: An overview of the project was provided.
3. **FITS demonstration and instructions**: FITS was introduced, and instructions were provided on using the Conceptboard.
4. **Testing phase**: Participants had 15 minutes to explore FITS independently with minimal guidance, mimicking real-world use. Afterward, they were asked to upload and analyze the provided documents. They documented their thoughts on the Conceptboard.
5. **Feedback discussion**: A moderated feedback session based on the structured question set.

### *6.2 Results of the Focus Groups*

Overall, the feedback from both focus groups revealed a strong interest in the proposed tool, with participants acknowledging its potential benefits for consumers. However, they noted that the current version is rudimentary, yet complex, and thus unintuitive for end users.

**Design and Usability.** Participants suggested FITS would be more effective if targeted to general consumers rather than researchers. The design was viewed as "too scientific", which could make it less approachable for non-expert users. There was a consensus that FITS could benefit from being more user-friendly and efficient in presenting relevant information. While the color scheme was criticized, the overall interface was seen as clean. Many found the ability to upload their data useful, although some experienced difficulties. Displaying navigation paths was suggested to improve usability. Despite these points, some participants said that they would use the tool.

**Information Content and Presentation.** Although the content of the texts was positively viewed and considered well-founded, their complexity was one of the primary concerns. Participants suggested using shorter formats with bullet points and highlighting key information. They also suggested overall summaries per topic and brand. Some wished for a simpler reading level or German texts. The defined classes were generally perceived as very positive. The pie chart was seen as unclear, leading to suggestions for simpler graphical representations like bar plots. Additionally, users wanted legends and explanations to clarify the data visualizations. The data was generally perceived as credible due to the references after the text passages. Some participants expressed a desire for more transparency regarding data sources and reliability. Furthermore, they wished for more complete information about each brand in each category, since they often received "no results found" as a response.

**Additional Features.** Participants wanted quick, clear comparisons between brands. They requested a rating system like a traffic light model. A brand selection feature, a saved search history, and better filtering options were requested for the search functionality. Suggestions included creating user accounts, adding export functions for sharing analyses, adding automatic translations, and providing FAQs. Another suggestion was to integrate FITS into existing tools, such as Statista.[15] Furthermore, the idea of a mobile app was expressed for easier access and navigation. Analyzing data from a screenshot or a link to a website was also proposed. Finally, some minor bugs were noted, which will not be elaborated on further.

**Discussion of the Results.** The feedback indicates a strong interest in FITS but highlights several areas for improvement. Much of the criticism concerns usability, but a market-ready tool was not the focus. We wanted to investigate whether such a tool is possible, what problems we would encounter, and how this kind of tool would be perceived. The criticism regarding the complexity and amount of data reflects the trade-off between credibility (R1) and accessibility. Since FITS prioritizes verified information from scientific and NGO sources, simplification poses a challenge. Future work could investigate automated summarization using large language models, though risks such as hallucination remain [13]. The idea of structuring the information better might be a valid solution. The desire for completeness, i.e., the system shows "no results found," points to limitations regarding the amount of data. Implementing rating systems poses methodological and ethical challenges. Existing indexes (e.g., Fashion Transparency Index) are limited as they are complex, only cover specific aspects, and are not user-friendly. Deciding which sustainability issues should be included and how to weigh them requires more research. In addition, quantifying some issues and providing that data are challenges we cannot address in this paper.

## 7 Discussion and Conclusion

This study introduced a novel approach to improving transparency in sustainability within the fashion industry by applying NLP. Specifically, we presented FITS, a tool built upon transformer-based models and a curated dataset (SustainableTextileCorpus) derived from credible sources. The research contributes to the growing field of AI for social good by addressing key challenges in data scarcity, data credibility, and consumer information access. Moreover, FITS is an example of a use case where domain-specific NLP remains relevant. In conclusion, this work illustrates the transformative role that AI can play in promoting climate change action.

[15] https://de.statista.com/

## 7.1 Summary and Interpretation of the Results

In the first part, we created a dataset from scientific publications and NGO reports as credible sources. The dataset was filtered with relevant keywords, split into shorter texts, and labeled with 19 sustainability classes. The final dataset, the SustainableTextileCorpus, contained 5093 examples, of which 582 were labeled. Nonetheless, the corpus is too small, imbalanced, and strongly biased for more common brands. This was visible in model training as the best model only achieved an average weighted F1-score of 0.683. Additionally, within the focus groups, the lack of data was also criticized. However, at the moment, there is not enough credible data about the sustainability of fashion brands. We hope that the amount of reliable data in this domain will increase in the future.

In the next part, we created a multi-label classification approach by fine-tuning transformers. We compared several BERT-based models against three baselines and used HPO to improve the performance. Keyword matching and the statistical baseline show significantly worse results, emphasizing the advantages of fine-tuning pre-trained language models. The tuned RoBERTa model performed best, with a weighted F1-score of 0.683 and a macro-F1-score of 0.479. The boost in performance with HPO over the non-tuned baseline demonstrates its value even with limited, imbalanced data. ClimateBERT and SciBERT unexpectedly performed worse than BERT despite the relatedness of the pretraining texts to the SustainableTextileCorpus.

Subsequently, we created FITS, an interactive tool to inform consumers about the sustainability of textile brands. The focus groups confirmed the demand for such tools and their potential to influence sustainable consumption, but also highlighted the need for data completeness, content clarity, and better visualizations. Five of the requirements for FITS were met: R1 (the data sources are peer-reviewed publications and NGO reports), R4 (it is possible to upload your own data), R5 (it sorts information into categories), R6 (it is possible to filter by brands), and R7 (it is interactive). Since the comprehensibility (R2) and information density (R3) were criticized, these requirements were not met. Participants complained about the usability, so it is not easily manageable (R8). Despite these limitations, FITS represents a prototype that informs consumers about the ecological and social sustainability of textile brands.

## 7.2 Broader Implications

This work shows the potential of ML to support climate change mitigation by increasing transparency in sustainability. The fashion industry is one of the most significant contributors to environmental pollution [7, 8] and lacks accessible, structured sustainability information [21, 14, 12], despite growing consumer interest in sustainable products [4]. Tools that simplify complex information can play a pivotal role in encouraging informed choices and more sustainable behaviors. In addition, consumer empowerment can pressure the fashion industry to adopt more environmentally and socially responsible practices [6]. Moreover, the findings highlight the effectiveness of domain-specific transformer models as both a technically robust and environmentally sustainable alternative to large language models, particularly in fact-critical areas like sustainability reporting.

The project tackles the lack of credible sustainability data and transparency in complex textile supply chains [9]. Without reliable data, consumers struggle to make informed decisions [6]. Therefore, FITS only uses credible NGO and research data.

This work also demonstrates the potential of AI in climate action. AI can handle complex sustainability challenges by processing large amounts of unstructured data to extract valuable insights. While this study focuses on the fashion industry, the methodology could be extended to other high-emission sectors, such as the production of food or electronic devices, where sustainability transparency is also critical, and consumers can make a difference. We also want to encourage interdisciplinary work with experts in both fields.

## 7.3 Limitations and Future Work

The most significant limitation is the lack of credible sustainability data about the fashion industry. Data sparsity affects model performance and the usefulness of FITS. Our data collection method might also take part in it. While it increases relevancy, important data can be missed, or false positives can occur. Furthermore, each instance was only labeled by one person, so that no inter-rater reliability could be calculated. A deeper issue lies in the fashion industry itself, where sustainability reporting is inconsistent, voluntary, and not standardized. Additionally, relying on existing reports limits the scope to well-documented brands and neglects fast-changing data. The ML algorithm is limited because transformers can only handle a fixed context length, so the texts must be cut off at a certain length.

The tool has several limitations. Due to the nature of its sources, its language is scientific and does not target end-consumers. Additionally, the format of three sentences keeps the information short but still includes noise. Since context is missing, the information could be misinterpreted or irrelevant. In addition, the functionality is somewhat limited, e.g., there is only one type of visualization. The focus group evaluation is limited since we only asked 20 people, and the group discussion format can influence individuals' opinions through the views of others. Additionally, the lack of usability hindered the unbiased evaluation of just the functionality and the idea of FITS.

Future work should focus on expanding the SustainableTextileCorpus by allowing for other data sources or improving the search for data. For instance, using Named Entity Recognition to aid the keyword search or a transformer model that considers the context and the meaning of the keywords. A daily data collection, e.g., via web-scraping, with continual learning of the ML model, could be implemented to keep the dataset up to date. Furthermore, the algorithm could be improved by language generation methods if they avoid hallucinations, e.g., with a fact-checking mechanism, generation could summarize the data, translate it, and adapt the style toward the target group. Future work for the tool development is discussed in Section 6.2. Addressing linguistic and cultural differences is another interesting extension of this work. All in all, we demonstrated a prototype of a fashion information tool for sustainability. Although there are still limitations, the first results are promising and could be further advanced in future work.

## Acknowledgements

All authors acknowledge financing by the ZuSiNa project (no. 67KI21009A) funded by the German Federal Ministry of the Environment, Nature Conservation, Nuclear Safety and Consumer Protection. This work was also partially funded by the Ministry of Science and Culture, Lower Saxony, Germany, through funds from the zukunft.niedersachsen (ZN3480).